\documentclass[twoside,11pt]{article}

\usepackage{automl2019}
\usepackage{xcolor}
\usepackage{makecell}
\usepackage{graphicx} 
\usepackage{epstopdf}
\DeclareGraphicsExtensions{.eps}

\usepackage{multirow}
\usepackage{algorithmic}
\usepackage{algorithm}
\usepackage{amsmath} 
\usepackage{amssymb}
\usepackage{amsxtra}
\usepackage{url} 
\usepackage{color} 
\usepackage{bm}
\usepackage{placeins}
\usepackage[caption=false]{subfig}
\usepackage{xspace}
\usepackage{rotating}
\usepackage{siunitx}
\usepackage{textcomp}%
\usepackage{algorithm}
\usepackage{algorithmic}

\newcommand{\wrt}{\text{w.r.t.}\xspace}

\newcommand{\myvector}[1]{\bm{#1}}
\newcommand{\myvec}[1]{\myvector{#1}}

\newcommand{\R}[1]{\mathbb{R}^{#1}}

\newcommand{\dist}[1]{\SI{#1}{\meter}}

\newcommand{\excise}[1]{}

\newcommand{\ui}[1]{a}

\newcommand{\xdash}[1][black]{\textcolor{#1}{\rule[0.5ex]{2em}{1.5pt}}}
\newcommand{\cbox}[1]{\raisebox{\depth}{\fcolorbox{#1}{#1}{\null}}}
\jmlrheading{Faust, Francis and Mehta}

\ShortHeadings{Evolving Rewards to Automate Reinforcement Learning}{Faust, Francis and Mehta}
\firstpageno{1}

\begin{document}

\title{Evolving Rewards to Automate Reinforcement Learning}

\author{\name Aleksandra Faust \email faust@google.com \\
        \name Anthony Francis \email centaur@google.com \\
        \name Dar Mehta \email darm@google.com \\
        \addr Robotics at Google, Mountain View, CA, 94043, USA}

\maketitle

\begin{abstract}
Many continuous control tasks have easily formulated objectives, yet using them directly as a reward in reinforcement learning (RL) leads to suboptimal policies. Therefore, many classical control tasks guide RL training using complex rewards, which require tedious hand-tuning. We automate the reward search with AutoRL, an evolutionary layer over standard RL that treats reward tuning as hyperparameter optimization and trains a population of RL agents to find a reward that maximizes the task objective. AutoRL, evaluated on four Mujoco continuous control tasks over two RL algorithms, shows improvements over baselines, with the the biggest uplift for more complex tasks. The video can be found at: \url{https://youtu.be/svdaOFfQyC8}.
\end{abstract}

\section{Introduction}
\label{sec:intro}
Despite solving a number of challenging problems \citep{q-opt,chen2015deepdriving,levine2016end}, training RL agents remains difficult and tedious. One culprit is reward design, which currently requires many iterations of manual tuning.
Reward is a scalar that communicates the task objective and desirable behaviors. Ideally, it should be an indicator of task completion \citep{sutton1992reinforcement} or a simple metric over resulting trajectories. 
Consider a Humanoid task \citep{tassa2012humanoid}, where objectives might be for the agent to travel as far or as fast as possible. Learning directly on these metrics is problematic, especially for high dimensional continuous control problems, for two reasons. First, RL requires the agent to explore until it stumbles onto the goal or improves the task metric, which can take a prohibitively long time \citep{DBLP:journals/corr/AndrychowiczWRS17}. Second, there are many ways to accomplish a task, and some are less than ideal. For example, the Humanoid might run while flailing its arms, or roll on the ground. Practitioners circumvent those cases through \textit{reward shaping} \citep{reward_shaping_definition}. Beginning with simple rewards, such as distance travelled, they train a policy and observe the training outcome. Next, practitioners either add terms to the reward that provide informative feedback about progress, such as energy spent or torso verticality, or tune the weights between the reward terms. Then they retrain the policy, observe, and repeat until training is tractable and the agent is well-behaved. In the Humanoid task, the agent is expected to maximize the speed and time alive while minimizing the control and impact cost. The standard reward collapses this multi-objective into one scalar with carefully selected weights. This human-intensive process raises questions: a) Can we automate the tuning and learn a proxy reward that both promotes the learning and meets the task objective? and b) Given an already hand-tuned reward, is there a better parameterization that accomplishes the same multi-objective?

Our prior work \citep{autorl} introduces AutoRL over Deep Deterministic Policy Gradients (DDPG, \cite{ddpg}) in the context of robot navigation to learn two end-to-end (sensor to controls) tasks: point-to-point and path-following. Using large scale hyper-parameter optimization, \cite{autorl} first find the reward parameters that maximize the goal reached sparse objective, and then find the neural network architecture that maximizes the learned proxy reward. 
In that setting, AutoRL improves both training stability and policy quality for sparse objectives for both tasks. But it remains an open question whether AutoRL, particularly the search for proxy rewards, is useful for other RL algorithms, tasks, and objectives.

This paper applies AutoRL's evolutionary reward search to four continuous control benchmarks from OpenAI Gym \citep{openaigym}, including Ant, Walker2D, HumanoidStandup, and Humanoid, over two RL algorithms: off-policy Soft Actor-Critic (SAC, \cite{DBLP:journals/corr/abs-1812-05905}) and on-policy Proximal Policy Optimization (PPO, \cite{DBLP:journals/corr/SchulmanWDRK17}). 
We optimize parameterized versions of the standard environment rewards (proxy rewards) over two different objectives: \textit{metric-based single-task objectives} including distance travelled and height reached, and the multi-objective \textit{standard returns} typically used in these environments. 
The results yield three findings. First, evolving rewards trains better policies than hand-tuned baselines, and on complex problems outperforms hyperparameter-tuned baselines, showing a 489\% gain over hyperparameter tuning on a single-task objective for SAC on the Humanoid task. Second, the optimization over simpler single-task objectives produces comparable results to the carefully hand-tuned standard returns, reducing the need for manual tuning of multi-objective tasks. Lastly, under the same training budget the reward tuning produces higher quality policies faster than tuning the learning hyperparameters.

\section{Related Work}
\label{sec:relwork}
\textit{AutoML and RL:} AutoML automates neural network architecture searches for supervised learning with RL \citep{automl-rl,automl-rl-2,cai-aaai,zoph-automl} and Evolutionary Algorithms (EA) \citep{real-automl-evol,real-image, liu-progressive-evol}, with the latter showing an edge \citep{evol_vs_rl}.
While RL is part of the AutoML toolset, tuning RL itself has been very limited. For example, EA mutated the actor network weights \citep{ga-rl}. 

\textit{Reward design:} Aside from reward shaping \citep{automl-rl-2}, reward design methods include curriculum learning \citep{reverse-currriculum,back-curriculum, gur2018learning}, bootstrapping \citep{residual-policy}, and Inverse Reinforcement Learning (IRL) \citep{ng-inverse-00}. AutoRL keeps task difficulty constant, trains policies from scratch, and requires no demonstrations, but it could be used in addition to curriculum and bootstrapping.

\textit{Large-scale hyperparameter optimization} has become a standard technique for improving deep RL performance by tuning learning hyperparameters \citep{follownet,gur2018learning}. 
AutoRL uses the same toolset but for reward and network tuning. \citep{autorl} uses sparse objectives, such as reaching a goal for point-to-point navigation, or waypoints achieved for a path-following task. This is effective because agents can be given a variety of tasks, many of which are easy to achieve on a true sparse objective (such as nearby goals in uncluttered environments). Here, we focus on metric-based objectives. 

\section{AutoRL}
\label{sec:methods}
\textit{Preliminaries:} Consider a Partially-Observable Markov Decision Process (POMDP) with continuous actions, $\mathcal{M}(S, O, A, \myvec{D}, R, \gamma)$. $S$ are states, $O \subset \R{d_o}$ are observations, $A \subset \R{d_a}$ are actions, and the system's unknown dynamics are modeled as a transition distribution $\myvec{D}: S \times A \times S \rightarrow [0,1].$ The reward, $R: S \times A \rightarrow \R{}$ is an immediate feedback to the RL agent, while $0 < \gamma \leq 1$ is a discount factor. The goal of RL is to find a policy, $\myvec{\hat{\pi}}_R: S \times A \rightarrow [0,1]$ that maximizes the expected cumulative discounted reward $R$ of trajectories $\mathcal{T}$ drawn from a set of initial conditions, $N \subset O,$ guided by the policy $\pi$ \wrt the system dynamics $\myvec{D}$:
\begin{equation}
\label{eq:return}
\myvec{\hat{\pi}}_R = \arg \max_{\myvec{\pi}} \mathbb{E}_{\mathcal{T} \sim (\myvec{D}, N, \myvec{\pi})}[\sum_{i=0}^{\|\mathcal{T}\|}\gamma^{i}R(\myvec{s_i},\myvec{a_i})].
\end{equation}
POMDPs model the world, observations, actions, and reward; all affect learning and performance, and are usually chosen through trial and error. 
AutoRL addresses the gap by applying the AutoML toolset to RL's pain points of POMDP modelling. 

\textit{Problem formulation:} An agent's success at a task can often be evaluated over a trajectory $\mathcal{T}$ with a metric $G(\mathcal{T})$. For example, success for the Humanoid Standup task is standing as tall as possible, while for Ant or Humanoid it is traveling as fast as possible. We can evaluate $G(\mathcal{T})$ over trajectories $\mathcal{T}$ drawn from initial conditions $N$ and controlled by a policy $\myvec{\pi}$ to gauge the policy's quality $J(\myvec{\pi})$ \wrt the fitness metric:
\begin{equation}
\label{eq:obj}
J(\myvec{\pi}) = \mathbb{E}_{\mathcal{T} \sim (\myvec{D}, N, \myvec{\pi})}[G(\mathcal{T})].
\end{equation}
While the reward $R$ measures immediate feedback to the agent, the objective metric $G$ reflects human-interpretable trajectory quality and induces an ordering over policies. Humans want high-quality trajectories, which are sparse, but RL agents learn best from dense feedback, so we use $G$ to help pick $R$ as follows.

Consider an augmented POMDP, $\tilde{\mathcal{M}}(S, O, A, \myvec{D}, R(\myvec{\theta}), \gamma, G),$ where $\myvec{\theta} \in \Theta$ is the parameter of a proxy reward function $R(\myvec{\theta})$. The goal of AutoRL is to solve $\tilde{\mathcal{M}}$ by finding a policy $\myvec{\tilde{\pi}}$ that maximizes the fitness metric $J$ defined in \eqref{eq:obj} given a population of agents $\myvec{\hat\pi}(R(\myvec{\theta}))$ with parameterization $\myvec{\theta}$ drawn from $\Theta$:
\begin{equation}
\label{eq:obj2}
\myvec{\tilde{\pi}}= \text{argmax}_{\{\myvec{\hat\pi}_{R(\myvec{\theta})} | \theta \sim \Theta\}} J(\myvec{\hat\pi}_{R(\myvec{\theta})}),\,\text{ where } \myvec{\hat\pi}_{R(\myvec{\theta})} \text{ is given in \eqref{eq:return}.}
\end{equation}
Proxy rewards, $R(\myvec{\theta}),$ may contain new features that guide learning, or may reweight the objective $G(\mathcal{T})$ to more closely match the gradient of the value function.

\textit{Algorithm:} Denote a policy $\myvec{\pi}(\myvec{\theta})$ learned with an externally-provided RL algorithm that optimizes the cumulative return \eqref{eq:return} for a fixed $\myvec{\theta}$ as:
\begin{equation}
\label{eq:rl}
\myvec{\pi}(\myvec{\theta}) = \text{RL}(R(\myvec{\theta}))
\end{equation}
and let $n_{g}$ be the maximum population size, with $n_{mc}$ as the number of parallel trials. 

We train a population of $n_{mc}$ parallel RL agents according to \eqref{eq:rl}, each initialized with a different reward parameterization $\myvec{\theta_i}, i \in \{1,\cdots, n_{g}\}.$ The parameterization for the first $n_{mc}$ agents are selected randomly. 
When training of the $i^{th}$ RL agent completes, Monte Carlo rollouts estimate the fitness metric \eqref{eq:obj} of the resulting policy $\myvec{\pi}(\myvec{\theta_i})$ to obtain, $j_i.$ This estimate and reward parameterization, $(\myvec{\theta_i}, j_i),$ are added to the population experience set $\Theta,$ while the policy and estimate, $(\myvec{\pi(\theta_i}), j_i),$ are added to the policy evaluation set, $\Pi.$ 
As the evaluation set $\Pi$ is updated, AutoRL continually updates the current best policy $\myvec{\tilde{\pi}}$ according to \eqref{eq:obj}. Next, AutoRL selects the parameterization for the next trial from the population experience, $\Theta,$ according to Gaussian Process Bandits \citep{gp-bandits}, and starts training a new RL agent. Finally, training stops after $n_{g}$ trials.    

AutoRL scales linearly with the population size $n_g$. Concurrent trials $n_{mc}$ must be smaller than $n_g,$ in order to have enough completed experience to select the next set of parameters. Overall, AutoRL requires $O(n_{mc})$ processors, and runs $\frac{n_g}{n_{mc}}$ times longer than vanilla RL.


\definecolor{green}{HTML}{2CA02C}
\definecolor{blue}{HTML}{1F77C4}
\definecolor{orange}{HTML}{FF7F0E}
\begin{figure*}[t]
	\begin{center}
		\begin{tabular}{ccccc}
		&\small Ant-v1 &\small  Walker2D-v1 &\small HumanoidStandup-v1 & \small Humanoid-v1 \\
            \multirow{2}{*}{\rotatebox{90}{\small Task Objective (meters)}}
			&
            \subfloat[\scriptsize SAC for Distance]{\includegraphics[trim=9mm 3mm 4mm
			3mm,clip,width=0.22\textwidth,height=2.2cm,keepaspectratio=false]{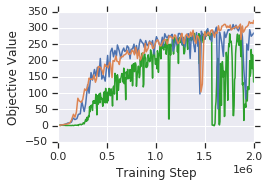}\label{fig:sac_ant_dg}} 
			&
			\subfloat[\scriptsize SAC for Distance]{\includegraphics[trim=9mm 3mm 4mm
			3mm,clip,width=0.22\textwidth,height=2.2cm,keepaspectratio=false]{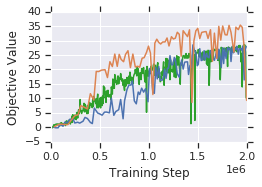}\label{fig:sac_walker2d_dg}} 
			&
			\subfloat[\scriptsize SAC for Height]{\includegraphics[trim=9mm 3mm 4mm
			3mm,clip,width=0.22\textwidth,height=2.2cm,keepaspectratio=false]{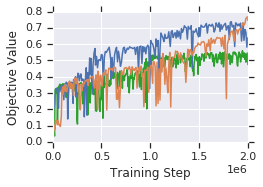}\label{fig:sac_standup_dg}} 
			&
			\subfloat[\scriptsize SAC for Distance]{\includegraphics[trim=9mm 3mm 4mm
			3mm,clip,width=0.22\textwidth,height=2.2cm,keepaspectratio=false]{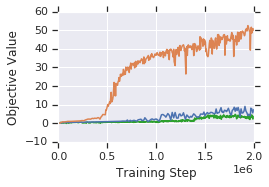}\label{fig:sac_humanoid_dg}}
			\\
			&
			\subfloat[\scriptsize PPO for Distance]{\includegraphics[trim=9mm 3mm 4mm
			3mm,clip,width=0.22\textwidth,height=2.2cm,keepaspectratio=false]{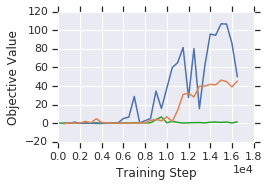}\label{fig:ppo_ant_dg}} 
            &
            \subfloat[\scriptsize PPO for Distance]{\includegraphics[trim=9mm 3mm 4mm
			3mm,clip,width=0.22\textwidth,height=2.2cm,keepaspectratio=false]{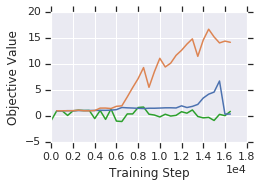}\label{fig:ppo_walker2d_dg}} 
            &
            \subfloat[\scriptsize PPO for Height]{\includegraphics[trim=9mm 3mm 4mm
			3mm,clip,width=0.22\textwidth,height=2.2cm,keepaspectratio=false]{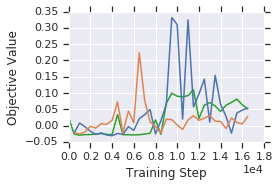}\label{fig:ppo_standup_dg}} 
            &
            \subfloat[\scriptsize PPO for Distance]{\includegraphics[trim=9mm 3mm 4mm
			3mm,clip,width=0.22\textwidth,height=2.2cm,keepaspectratio=false]{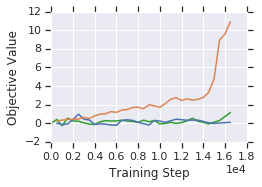}\label{fig:ppo_humanoid_dg}}
            \\
		    \multirow{2}{*}{\rotatebox{90}{\small Std. Objective}}
            &
            \subfloat[\scriptsize SAC / Std. Reward]{\includegraphics[trim=9mm 3mm 4mm
			3mm,clip,width=0.22\textwidth,height=2.2cm,keepaspectratio=false]{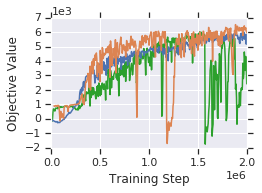}\label{fig:sac_ant_sr}} 
			&
			\subfloat[\scriptsize SAC / Std. Reward]{\includegraphics[trim=9mm 3mm 4mm
			3mm,clip,width=0.22\textwidth,height=2.2cm,keepaspectratio=false]{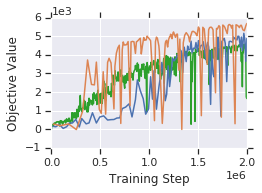}\label{fig:sac_walker_sr}} 
			&
			\subfloat[\scriptsize SAC / Std. Reward]{\includegraphics[trim=9mm 3mm 4mm
			3mm,clip,width=0.22\textwidth,height=2.2cm,keepaspectratio=false]{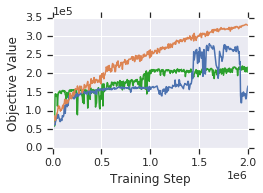}\label{fig:sac_standup_sr}} 
			&
			\subfloat[\scriptsize SAC / Std. Reward]{\includegraphics[trim=9mm 3mm 4mm
			3mm,clip,width=0.22\textwidth,height=2.2cm,keepaspectratio=false]{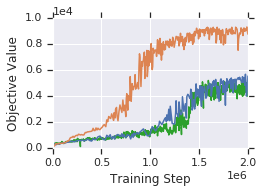}\label{fig:sac_humanoid_sr}}
			\\
			&
            \subfloat[\scriptsize PPO / Std. Reward]{\includegraphics[trim=9mm 3mm 4mm
			3mm,clip,width=0.22\textwidth,height=2.2cm,keepaspectratio=false]{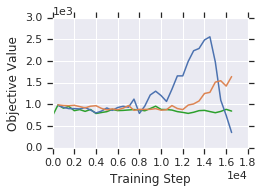}\label{fig:ppo_ant_sr}} 
            &
            \subfloat[\scriptsize PPO / Std. Reward]{\includegraphics[trim=9mm 3mm 4mm
			3mm,clip,width=0.22\textwidth,height=2.2cm,keepaspectratio=false]{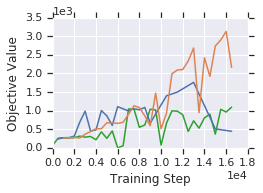}\label{fig:ppo_walker_sr}} 
            &
            \subfloat[\scriptsize PPO / Std. Reward]{\includegraphics[trim=9mm 3mm 4mm
			3mm,clip,width=0.22\textwidth,height=2.2cm,keepaspectratio=false]{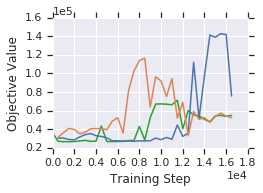}\label{fig:ppo_standup_sr}} 
            &
            \subfloat[\scriptsize PPO / Std. Reward]{\includegraphics[trim=9mm 3mm 4mm
			3mm,clip,width=0.22\textwidth,height=2.2cm,keepaspectratio=false]{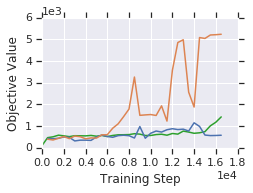}\label{fig:ppo_humanoid_sr}} 
		\end{tabular}
		\begin{tabular}{ccc}
       \xdash[green]\scriptsize Baseline (HT) \citep{openaigym} & \xdash[blue]\scriptsize Hyperparameter (HP) Tuned Baseline & \xdash[orange]\scriptsize AutoRL \textbf{(ours)}
        \end{tabular}
		\caption{\small Task objective results a-d) using SAC and e-h) using PPO and  standard return objective i-l) using SAC and m-p) using PPO for Ant, Walker2d, Humanoid Standup, and Humanoid. \label{fig:sac_ppo_dg_evalOnDG} \label{fig:sac_ppo_sr_evalOnSR}}
	\end{center}
\end{figure*}

\section{Results}
\label{sec:results}
We implement AutoRL hyperparameter optimization with Vizier \citep{vizier} over the PPO and SAC and apply it to four widely-used MuJoCo \citep{mujoco} continuous control environments in order of increasing complexity: Ant, Walker, Humanoid Standup, and Humanoid (Table \ref{tab:environments}; see Appendix \ref{sec:params}). 

To assess AutoRL's ability to reduce reward engineering while maintaining quality on existing metrics, we contrast two objectives: task objectives and standard returns. \textit{Task objectives} measure task achievement for continuous control: distance traveled for Ant, Walker, and Humanoid, and height achieved for Standup. \textit{Standard returns} are the metrics by which tasks are normally evaluated. 
For both objectives, the parameterized reward $\myvec{\theta}$ is a re-weighted standard reward (see Appendix \ref{sec:sparse}).  
We compare AutoRL with two baselines, hand-tuned and hyperparameter-tuned. Hand-tuned (HT) uses default learning parameters for each algorithm. Hyperparameter-tuned (HP) uses Vizier to optimize learning hyperparameters such as learning rate and discount factor. In all cases, AutoRL uses HT's default hyperparameters. 
We train up to $n_{g} = 1000$ agents parallelized across $n_{mc} = 100$ workers. SAC trains for 2 million steps, while PPO's training episodes depends on the environment (Table \ref{tab:environments}; see Appendix \ref{sec:params}). Policy quality (fitness metric \eqref{eq:obj}) \wrt the objective is evaluated over 50 trajectories.

\textit{Task Objective Evaluation:} AutoRL outperforms the HP tuned baseline (blue) for all tasks trained with SAC (Figures \ref{fig:sac_ppo_dg_evalOnDG}a-d) and on Walker and Humanoid for PPO (Figures \ref{fig:sac_ppo_dg_evalOnDG}e-h). Both outperform the hand-tuned baseline, whose parameters AutoRL uses. Note that AutoRL uses non-tuned learning hyperparameters. AutoRL's benefit over HP tuning is consistent - though relatively small for simpler tasks - but is very noticeable on the most complex task, Humanoid, with 489\% improvement over HP tuning.  AutoRL shows 64\% improvement on Humanoid. It is interesting to note that in some cases AutoRL converges more slowly, and does not reach peak performance until very late in the training process (Figures \ref{fig:sac_ant_dg}, \ref{fig:sac_standup_dg}, and \ref{fig:ppo_humanoid_dg}). We suspect this is because the tasks simply require more training iterations to converge, and the baselines end up getting stuck in a local minima, while AutoRL manages to escape it.

\textit{Std. Reward Evaluation:} AutoRL outperforms HP tuning on all SAC tasks (Figure \ref{fig:sac_ppo_sr_evalOnSR}i-l) and on Walker and Humanoid for PPO (Figure \ref{fig:sac_ppo_sr_evalOnSR}m-p). Both beat the hand-tuned baseline.

\textit{Single-Task vs Multi-Objective:} If our goal is finding objectives that are easier to provide than hand-tuning the multiple objectives of a standard reward, then we want to know how well AutoRL optimizes simple task objectives. On  Humanoid, the task objective agents travel the farthest for both SAC and PPO (Figures \ref{fig:sac_sr_dg_on_dg} and \ref{fig:ppo_sr_dg_on_dg} dark red vs. light red), while on other tasks optimizing over the task objective is comparable to optimizing over the standard reward. Task objectives and standard returns have similar performance, suggesting task objectives obtain good policies with reduced reward engineering effort (see Appendix \ref{sec:cross-eval} for further discussion). Unsurprisingly, AutoRL over the standard reward produces the highest scores, when evaluated on the standard reward in 13 out of 16 conditions (Walker, Standup in Figure \ref{fig:sac_sr_dg_on_sr}, and Ant in Figure \ref{fig:ppo_sr_dg_on_sr}. However, videos show the policies differ in style: Humanoid optimized for the standard reward produces a jumping and falling loop, while the height-reached policy stably rises to just above kneeling.\footnote{\url{https://youtu.be/svdaOFfQyC8}} We leave it for future work to apply AutoRL for non-scalarized multi-objective optimization problems.

\definecolor{sr_hp_light_blue}{HTML}{6D9EEB}
\definecolor{dg_hp_dark_blue}{HTML}{1155CC}
\definecolor{sr_r_light_red}{HTML}{E06666}
\definecolor{dg_r_dark_red}{HTML}{990000}
\begin{figure*}[t]
	\begin{center}
		\begin{tabular}{cccc}
		\rotatebox{90}{\small Task Objective}
			 &\subfloat[SAC on Task obj.]{\includegraphics[trim=7mm 5mm 22mm 7mm,clip,width=0.3\textwidth,height=3.0cm,keepaspectratio=false]{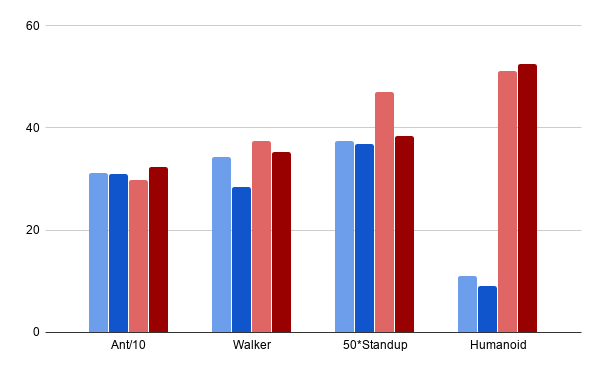}\label{fig:sac_sr_dg_on_dg}} 
			&
			\subfloat[PPO on Task. obj.]{\includegraphics[trim=7mm 5mm 22mm 7mm,clip,width=0.3\textwidth,height=3.0cm,keepaspectratio=false]{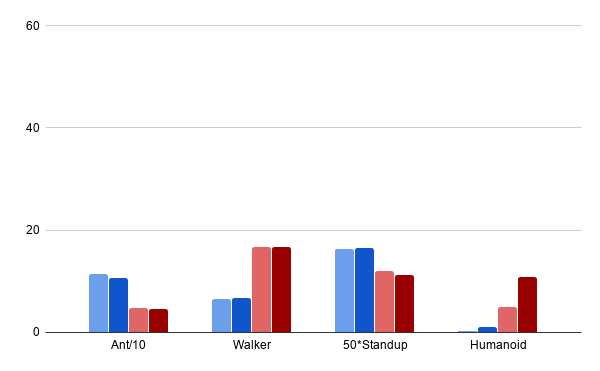}\label{fig:ppo_sr_dg_on_dg}} 
			& \subfloat[Std. Return SAC Humanoid]{\includegraphics[trim=0mm 0mm 0mm 0mm,clip,width=0.3\textwidth,height=3.0cm,keepaspectratio=false]{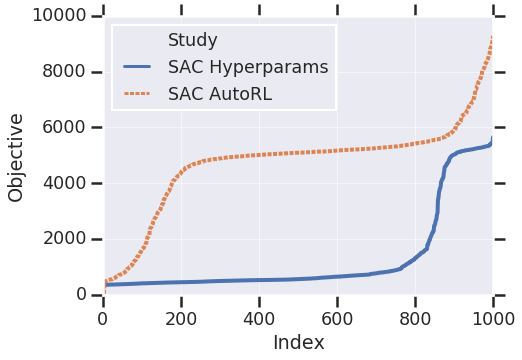}\label{fig:task_params_vs_rewards_sac}}\\
		 \rotatebox{90}{\small Std. reward Obj.}
            &\subfloat[SAC on Std. Reward]{\includegraphics[trim=7mm 5mm 22mm 7mm,clip,width=0.3\textwidth,height=3.0cm,keepaspectratio=false]{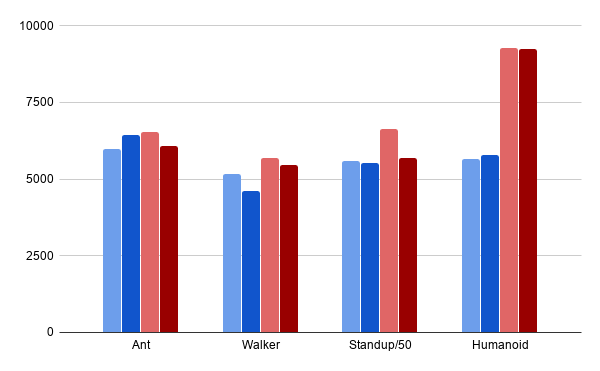}\label{fig:sac_sr_dg_on_sr}} 
			&
			\subfloat[PPO on Std. Reward]{\includegraphics[trim=7mm 5mm 22mm 7mm,clip,width=0.3\textwidth,height=3.0cm,keepaspectratio=false]{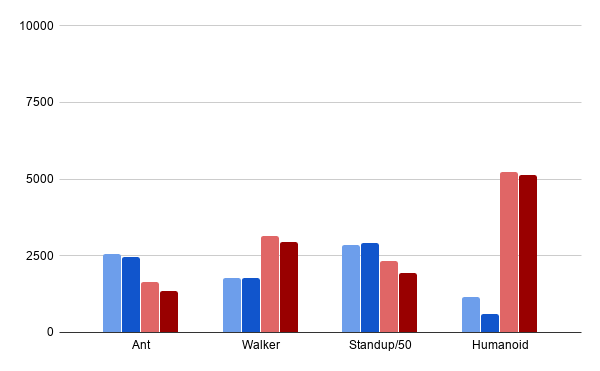}\label{fig:ppo_sr_dg_on_sr}}
			&
			\subfloat[Std. Return PPO Humanoid]{\includegraphics[trim=0mm 0mm 0mm
			0mm,clip,width=0.3\textwidth,height=3.0cm,keepaspectratio=false]{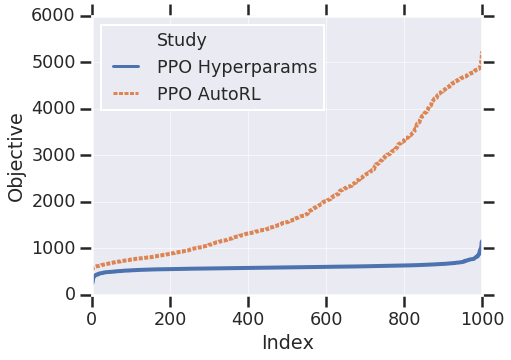}\label{fig:task_params_vs_rewards_ppo}}
			\\
			 
		\end{tabular}
		\begin{tabular}{cccc}
        \cbox{sr_hp_light_blue} \scriptsize Std. Ret. HP Tuned Baseline &\cbox{dg_hp_dark_blue} \scriptsize Task Obj. HP Tuned Baseline &
        \cbox{sr_r_light_red} \scriptsize Std. Ret. AutoRL & \cbox{dg_r_dark_red} \scriptsize Task Obj. AutoRL
        \end{tabular}
		\caption{\small Cross-evaluation \wrt standard returns (SAC (a), PPO (b)) and task objectives (SAC (d), PPO (e)). Standup and Ant are scaled.  
		c) and f) Sorted Vizier trials show the benefits of reward tuning over hyperparameter tuning.
		\label{fig:meta_objective_comparison}\label{fig:task_params_vs_rewards}}
	\end{center}
\end{figure*}
\label{sec:toe}

\textit{Reward vs HP tuning:} AutoRL shows promising benefits for hyperparameter optimization performance: evolving rewards produces more high-performing trials and has a higher peak performance than hyperparameter tuning on both SAC and PPO (Fig. \ref{fig:task_params_vs_rewards} a,b,d,e). Given a limited computational training budget, reward tuning explores better policies than hyperparameter tuning (Fig. \ref{fig:task_params_vs_rewards} c,f) and is more likely to produce good policies.

\subsection{Conclusion}
\label{sec:conclusion}
In this paper we learn proxy rewards for continuous control tasks with AutoRL, a method that automates RL reward design by using evolutionary optimization over a given objective. Benchmarking over two RL algorithms, four MujoCo tasks, and two different true objectives show that: a) AutoRL outperforms both hand-tuned and learning hyperparameter tuned RL; b) produces comparable and often superior policies with a simpler true objective, hence reducing human engineering time; and c) often produces better policies faster than hyperparameter tuning, suggesting that under a limited training budget tuning proxy rewards might be more beneficial than tuning hyperparameter and that a more in-depth analysis would be appropriate. All three conclusions hold even stronger for more complex environments such as Humanoid, making AutoRL a promising technique for training RL agents for complex tasks, with less hand engineering and better results.

\acks{\small We thank Oscar Ramirez, Rico Jonschkowski, Shane Gu, Sam Fishman, Eric Jang, Sergio Guadarrama, Sergey Levine, Brian Ichter, Hao-Tien Chiang, Jie Tan \& Vincent Vanhoucke for their input.} 

\bibliography{literature}

\begin{thebibliography}{34}
\providecommand{\natexlab}[1]{#1}
\providecommand{\url}[1]{\texttt{#1}}
\expandafter\ifx\csname urlstyle\endcsname\relax
  \providecommand{\doi}[1]{doi: #1}\else
  \providecommand{\doi}{doi: \begingroup \urlstyle{rm}\Url}\fi

\bibitem[Andrychowicz et~al.(2017)Andrychowicz, Wolski, Ray, Schneider, Fong,
  Welinder, McGrew, Tobin, Abbeel, and
  Zaremba]{DBLP:journals/corr/AndrychowiczWRS17}
Marcin Andrychowicz, Filip Wolski, Alex Ray, Jonas Schneider, Rachel Fong,
  Peter Welinder, Bob McGrew, Josh Tobin, Pieter Abbeel, and Wojciech Zaremba.
\newblock Hindsight experience replay.
\newblock \emph{CoRR}, abs/1707.01495, 2017.
\newblock URL \url{http://arxiv.org/abs/1707.01495}.

\bibitem[Brockman et~al.(2016)Brockman, Cheung, Pettersson, Schneider,
  Schulman, Tang, and Zaremba]{openaigym}
Greg Brockman, Vicki Cheung, Ludwig Pettersson, Jonas Schneider, John Schulman,
  Jie Tang, and Wojciech Zaremba.
\newblock Openai gym, 2016.

\bibitem[Cai et~al.(2018)Cai, Chen, Zhang, Yu, and Wang]{cai-aaai}
Han Cai, Tianyao Chen, Weinan Zhang, Yong Yu, and Jun Wang.
\newblock Efficient architecture search by network transformation.
\newblock In \emph{Proceedings of the Thirty-Second {AAAI} Conference on
  Artificial Intelligence, (AAAI-18), the 30th innovative Applications of
  Artificial Intelligence (IAAI-18), and the 8th {AAAI} Symposium on
  Educational Advances in Artificial Intelligence (EAAI-18), New Orleans,
  Louisiana, USA, February 2-7, 2018}, pages 2787--2794, 2018.

\bibitem[Chen et~al.(2015)Chen, Seff, Kornhauser, and
  Xiao]{chen2015deepdriving}
Chenyi Chen, Ari Seff, Alain Kornhauser, and Jianxiong Xiao.
\newblock Deepdriving: Learning affordance for direct perception in autonomous
  driving.
\newblock In \emph{Proceedings of the IEEE International Conference on Computer
  Vision}, pages 2722--2730, 2015.

\bibitem[Chiang et~al.(2019)Chiang, Faust, Fiser, and Francis]{autorl}
Hao-Tien~Lewis Chiang, Aleksandra Faust, Marek Fiser, and Anthony Francis.
\newblock Learning navigation behaviors end-to-end with autorl.
\newblock \emph{IEEE Robotics and Automation Letters}, 4\penalty0 (2):\penalty0
  2007--2014, April 2019.
\newblock ISSN 2377-3766.
\newblock \doi{10.1109/LRA.2019.2899918}.

\bibitem[Erez et~al.(2011)Erez, Tassa, and Todorov]{erez2011infinite}
Tom Erez, Yuval Tassa, and Emanuel Todorov.
\newblock Infinite horizon model predictive control for nonlinear periodic
  tasks.
\newblock \emph{Manuscript under review}, 4, 2011.

\bibitem[Florensa et~al.(2017)Florensa, Held, Wulfmeier, Zhang, and
  Abbeel]{reverse-currriculum}
Carlos Florensa, David Held, Markus Wulfmeier, Michael Zhang, and Pieter
  Abbeel.
\newblock Reverse curriculum generation for reinforcement learning.
\newblock In Sergey Levine, Vincent Vanhoucke, and Ken Goldberg, editors,
  \emph{Proceedings of the 1st Annual Conference on Robot Learning}, volume~78
  of \emph{Proceedings of Machine Learning Research}, pages 482--495. PMLR,
  13--15 Nov 2017.

\bibitem[Golovin et~al.(2017)Golovin, Solnik, Moitra, Kochanski, Karro, and
  Sculley]{vizier}
Daniel Golovin, Benjamin Solnik, Subhodeep Moitra, Greg Kochanski, John Karro,
  and D.~Sculley.
\newblock Google vizier: A service for black-box optimization.
\newblock In \emph{Proc. of ACM International Conference on Knowledge Discovery
  and Data Mining}, pages 1487--1495. ACM, 2017.
\newblock \doi{10.1145/3097983.3098043}.

\bibitem[Guadarrama et~al.(2018)Guadarrama, Korattikara, Ramirez, Castro,
  Holly, Fishman, Wang, Gonina, Harris, Vanhoucke, and Brevdo]{tfagents}
Sergio Guadarrama, Anoop Korattikara, Oscar Ramirez, Pablo Castro, Ethan Holly,
  Sam Fishman, Ke~Wang, Ekaterina Gonina, Chris Harris, Vincent Vanhoucke, and
  Eugene Brevdo.
\newblock {TF-Agents}: A library for reinforcement learning in tensorflow.
\newblock \url{https://github.com/tensorflow/agents}, 2018.
\newblock URL \url{https://github.com/tensorflow/agents}.

\bibitem[Gur et~al.(2019)Gur, Rueckert, Faust, and
  Hakkani-Tur]{gur2018learning}
Izzeddin Gur, Ulrich Rueckert, Aleksandra Faust, and Dilek Hakkani-Tur.
\newblock Learning to navigate the web.
\newblock In \emph{International Conference on Learning Representations}, 2019.
\newblock URL \url{https://openreview.net/forum?id=BJemQ209FQ}.

\bibitem[Haarnoja et~al.(2018{\natexlab{a}})Haarnoja, Zhou, Ha, Tan, Tucker,
  and Levine]{ec-sac}
Tuomas Haarnoja, Aurick Zhou, Sehoon Ha, Jie Tan, George Tucker, and Sergey
  Levine.
\newblock Learning to walk via deep reinforcement learning.
\newblock \emph{CoRR}, abs/1812.11103, 2018{\natexlab{a}}.
\newblock URL \url{http://arxiv.org/abs/1812.11103}.

\bibitem[Haarnoja et~al.(2018{\natexlab{b}})Haarnoja, Zhou, Hartikainen,
  Tucker, Ha, Tan, Kumar, Zhu, Gupta, Abbeel, and
  Levine]{DBLP:journals/corr/abs-1812-05905}
Tuomas Haarnoja, Aurick Zhou, Kristian Hartikainen, George Tucker, Sehoon Ha,
  Jie Tan, Vikash Kumar, Henry Zhu, Abhishek Gupta, Pieter Abbeel, and Sergey
  Levine.
\newblock Soft actor-critic algorithms and applications.
\newblock \emph{CoRR}, abs/1812.05905, 2018{\natexlab{b}}.
\newblock URL \url{http://arxiv.org/abs/1812.05905}.

\bibitem[Ivanovic et~al.(2018)Ivanovic, Harrison, Sharma, Chen, and
  Pavone]{back-curriculum}
Boris Ivanovic, James Harrison, Apoorva Sharma, Mo~Chen, and Marco Pavone.
\newblock Barc: Backward reachability curriculum for robotic reinforcement
  learning.
\newblock \emph{CoRR}, abs/1806.06161, 2018.

\bibitem[Kalashnikov et~al.(2018)Kalashnikov, Irpan, Pastor, Ibarz, Herzog,
  Jang, Quillen, Holly, Kalakrishnan, Vanhoucke, and Levine]{q-opt}
Dmitry Kalashnikov, Alex Irpan, Peter Pastor, Julian Ibarz, Alexander Herzog,
  Eric Jang, Deirdre Quillen, Ethan Holly, Mrinal Kalakrishnan, Vincent
  Vanhoucke, and Sergey Levine.
\newblock Scalable deep reinforcement learning for vision-based robotic
  manipulation.
\newblock In Aude Billard, Anca Dragan, Jan Peters, and Jun Morimoto, editors,
  \emph{Proceedings of The 2nd Conference on Robot Learning}, volume~87 of
  \emph{Proceedings of Machine Learning Research}, pages 651--673. PMLR, 29--31
  Oct 2018.

\bibitem[Khadka and Tumer(2018)]{ga-rl}
Shauharda Khadka and Kagan Tumer.
\newblock Evolution-guided policy gradient in reinforcement learning.
\newblock In S.~Bengio, H.~Wallach, H.~Larochelle, K.~Grauman, N.~Cesa-Bianchi,
  and R.~Garnett, editors, \emph{Advances in Neural Information Processing
  Systems 31}, pages 1188--1200. Curran Associates, Inc., 2018.

\bibitem[Levine et~al.(2016)Levine, Finn, Darrell, and Abbeel]{levine2016end}
Sergey Levine, Chelsea Finn, Trevor Darrell, and Pieter Abbeel.
\newblock End-to-end training of deep visuomotor policies.
\newblock \emph{Journal of Machine Learning Research}, 17\penalty0
  (39):\penalty0 1--40, 2016.

\bibitem[Lillicrap et~al.(2015)Lillicrap, Hunt, Pritzel, Heess, Erez, Tassa,
  Silver, and Wierstra]{ddpg}
Timothy~P. Lillicrap, Jonathan~J. Hunt, Alexander Pritzel, Nicolas Heess, Tom
  Erez, Yuval Tassa, David Silver, and Daan Wierstra.
\newblock Continuous control with deep reinforcement learning.
\newblock \emph{CoRR}, abs/1509.02971, 2015.
\newblock URL \url{http://arxiv.org/abs/1509.02971}.

\bibitem[Liu et~al.(2017)Liu, Zoph, Shlens, Hua, Li, Fei{-}Fei, Yuille, Huang,
  and Murphy]{liu-progressive-evol}
Chenxi Liu, Barret Zoph, Jonathon Shlens, Wei Hua, Li{-}Jia Li, Li~Fei{-}Fei,
  Alan~L. Yuille, Jonathan Huang, and Kevin Murphy.
\newblock Progressive neural architecture search.
\newblock \emph{CoRR}, abs/1712.00559, 2017.
\newblock URL \url{http://arxiv.org/abs/1712.00559}.

\bibitem[Ng and Russell(2000)]{ng-inverse-00}
Andrew~Y. Ng and Stuart~J. Russell.
\newblock Algorithms for inverse reinforcement learning.
\newblock In \emph{Proceedings of the Seventeenth International Conference on
  Machine Learning}, ICML '00, pages 663--670, San Francisco, CA, USA, 2000.
  Morgan Kaufmann Publishers Inc.

\bibitem[Real et~al.(2017)Real, Moore, Selle, Saxena, Suematsu, Tan, Le, and
  Kurakin]{real-automl-evol}
Esteban Real, Sherry Moore, Andrew Selle, Saurabh Saxena, Yutaka~Leon Suematsu,
  Jie Tan, Quoc~V. Le, and Alexey Kurakin.
\newblock Large-scale evolution of image classifiers.
\newblock In \emph{Proceedings of the 34th International Conference on Machine
  Learning - Volume 70}, ICML'17, pages 2902--2911. JMLR.org, 2017.

\bibitem[Real et~al.(2018{\natexlab{a}})Real, Aggarwal, Huang, and
  Le.]{evol_vs_rl}
Esteban Real, Alok Aggarwal, Yanping Huang, and Quoc~V. Le.
\newblock Evolutionary algorithms and reinforcement learning: A comparative
  case study for architecture search.
\newblock In \emph{AutoML@ICMLRobotics and Automation, 2002. Proceedings.
  ICRA'02. IEEE International Conference on}, 2018{\natexlab{a}}.

\bibitem[Real et~al.(2018{\natexlab{b}})Real, Aggarwal, Huang, and
  Le]{real-image}
Esteban Real, Alok Aggarwal, Yanping Huang, and Quoc~V. Le.
\newblock Regularized evolution for image classifier architecture search.
\newblock \emph{CoRR}, abs/1802.01548, 2018{\natexlab{b}}.
\newblock URL \url{http://arxiv.org/abs/1802.01548}.

\bibitem[Schulman et~al.(2015)Schulman, Moritz, Levine, Jordan, and
  Abbeel]{Schulman2015HighDimensionalCC}
John Schulman, Philipp Moritz, Sergey Levine, Michael~I. Jordan, and Pieter
  Abbeel.
\newblock High-dimensional continuous control using generalized advantage
  estimation.
\newblock \emph{CoRR}, abs/1506.02438, 2015.

\bibitem[Schulman et~al.(2017)Schulman, Wolski, Dhariwal, Radford, and
  Klimov]{DBLP:journals/corr/SchulmanWDRK17}
John Schulman, Filip Wolski, Prafulla Dhariwal, Alec Radford, and Oleg Klimov.
\newblock Proximal policy optimization algorithms.
\newblock \emph{CoRR}, abs/1707.06347, 2017.
\newblock URL \url{http://arxiv.org/abs/1707.06347}.

\bibitem[Shah et~al.(2018)Shah, Fiser, Faust, Kew, and Hakkani-Tur]{follownet}
Pararth Shah, Marek Fiser, Aleksandra Faust, Chase Kew, and Dilek Hakkani-Tur.
\newblock Follownet: Robot navigation by following natural language directions
  with deep reinforcement learning.
\newblock In \emph{Third Machine Learning in Planning and Control of Robot
  Motion Workshop at ICRA}, 2018.

\bibitem[Silver et~al.(2018)Silver, Allen, Tenenbaum, and
  Kaelbling]{residual-policy}
Tom Silver, Kelsey Allen, Josh Tenenbaum, and Leslie~Pack Kaelbling.
\newblock Residual policy learning.
\newblock \emph{CoRR}, abs/1812.06298, 2018.
\newblock URL \url{http://arxiv.org/abs/1812.06298}.

\bibitem[Srinivas et~al.(2012)Srinivas, Krause, Kakade, and Seeger]{gp-bandits}
Niranjan Srinivas, Andreas Krause, Sham Kakade, and Matthias Seeger.
\newblock Information-theoretic regret bounds for gaussian process optimization
  in the bandit setting.
\newblock \emph{IEEE Transactions on Information Theory}, 58\penalty0
  (5):\penalty0 3250--3265, May 2012.
\newblock \doi{10.1109/TIT.2011.2182033}.

\bibitem[Sutton et~al.(1992)Sutton, Barto, and
  Williams]{sutton1992reinforcement}
Richard~S Sutton, Andrew~G Barto, and Ronald~J Williams.
\newblock Reinforcement learning is direct adaptive optimal control.
\newblock \emph{IEEE Control Systems}, 12\penalty0 (2):\penalty0 19--22, 1992.

\bibitem[{Tassa} et~al.(2012){Tassa}, {Erez}, and {Todorov}]{tassa2012humanoid}
Y.~{Tassa}, T.~{Erez}, and E.~{Todorov}.
\newblock Synthesis and stabilization of complex behaviors through online
  trajectory optimization.
\newblock In \emph{2012 IEEE/RSJ International Conference on Intelligent Robots
  and Systems}, pages 4906--4913, Oct 2012.
\newblock \doi{10.1109/IROS.2012.6386025}.

\bibitem[Todorov et~al.(2012)Todorov, Erez, and Tassa]{mujoco}
Emanuel Todorov, Tom Erez, and Yuval Tassa.
\newblock Mujoco: A physics engine for model-based control.
\newblock \emph{2012 IEEE/RSJ International Conference on Intelligent Robots
  and Systems}, pages 5026--5033, 2012.

\bibitem[Wiewiora(2010)]{reward_shaping_definition}
Eric Wiewiora.
\newblock \emph{Reward Shaping}, pages 863--865.
\newblock Springer US, Boston, MA, 2010.
\newblock ISBN 978-0-387-30164-8.
\newblock \doi{10.1007/978-0-387-30164-8_731}.
\newblock URL \url{https://doi.org/10.1007/978-0-387-30164-8_731}.

\bibitem[Zoph and Le(2017)]{automl-rl}
Barret Zoph and Quoc~V. Le.
\newblock Neural architecture search with reinforcement learning.
\newblock In \emph{5th International Conference on Learning Representations,
  {ICLR} 2017, Toulon, France, April 24-26, 2017, Conference Track
  Proceedings}, 2017.

\bibitem[Zoph et~al.(2018{\natexlab{a}})Zoph, Vasudevan, Shlens, and
  Le]{automl-rl-2}
Barret Zoph, Vijay Vasudevan, Jonathon Shlens, and Quoc~V. Le.
\newblock Learning transferable architectures for scalable image recognition.
\newblock In \emph{2018 {IEEE} Conference on Computer Vision and Pattern
  Recognition, {CVPR} 2018, Salt Lake City, UT, USA, June 18-22, 2018}, pages
  8697--8710, 2018{\natexlab{a}}.

\bibitem[Zoph et~al.(2018{\natexlab{b}})Zoph, Vasudevan, Shlens, and
  Le]{zoph-automl}
Barret Zoph, Vijay Vasudevan, Jonathon Shlens, and Quoc~V. Le.
\newblock Learning transferable architectures for scalable image recognition.
\newblock In \emph{2018 {IEEE} Conference on Computer Vision and Pattern
  Recognition, {CVPR} 2018, Salt Lake City, UT, USA, June 18-22, 2018}, pages
  8697--8710, 2018{\natexlab{b}}.

\end{thebibliography}

\newpage  
\appendix

\section{AutoRL Parameter Settings}
\label{sec:params}
We use the TF Agents \citep{tfagents} implementation of SAC \citep{DBLP:journals/corr/abs-1812-05905, ec-sac} and PPO \citep{DBLP:journals/corr/SchulmanWDRK17}. 
In the PPO implementation, training steps shown in the charts are equal to the number of gradient update steps, and training is done 25 times every N episodes where N is environment-specific and defined in Table \ref{tab:environments}. 
For Ant and Walker the approximate number of environment steps is 19.8 million, and for Standup and Humanoid it is 158.4 million.

\begin{table}[h]
\scriptsize
    \caption{Environments}
	\label{tab:environments}
    \centering
    \begin{tabular}{r|l|l|r|r|r}
        \hline \hline
        \textbf{Name} & \textbf{Description} & \textbf{Reference} & \textbf{PPO Episodes} & \textbf{State}& \textbf{Action} \\
         &  &  &  \textbf{per Iteration} & \textbf{Dim.}& \textbf{Dim.} \\\hline
        \textbf{Ant} & Ant 3D Locomotion & \cite{Schulman2015HighDimensionalCC} &  30 & 111 & 8\\
        \textbf{Walker} & Walker 2D Locomotion & \cite{erez2011infinite} & 30 &17 &6 \\
        \textbf{Standup} & Humanoid Standup & N/A & 240 &376 &17\\
        \textbf{Humanoid} & Humanoid Locomotion & \cite{tassa2012humanoid} & 240 & 376 &17\\ \hline \hline
    \end{tabular}
\end{table}

\begin{table}[h]
\scriptsize
    \caption{Tuned Reward Parameters}
	\label{tab:tuned_rewards}
    \centering
    \begin{tabular}{r|cccc}
        \hline \hline
        \textbf{Parameters} & \textbf{Ant}& \textbf{Walker}& \textbf{Standup}& \textbf{Humanoid} \\ \hline
        \textbf{Achievement} & Linear Velocity & Linear Velocity  & Height / Time & Linear Velocity  \\
        \textbf{Cost} & Control Cost & Control Cost & Quadratic Cost & Control Cost \\
        \textbf{Impact} & Contact Cost & N/A & Quadratic Impact & Impact Cost \\
        \textbf{Survival} & Alive Bonus & Alive Bonus& Alive Bonus& Alive Bonus \\ \hline \hline
    \end{tabular}
\end{table}
\setlength{\tabcolsep}{6pt}

\setlength{\tabcolsep}{5pt}
\begin{table}[tb]
\scriptsize
    \caption{Tuned Hyperparameters}
	\label{tab:tuned_hyperparameters}
    \centering
    \begin{tabular}{c|l|r|c|l|r}
        \hline \hline
        \textbf{Algorithm} & \textbf{Parameters} & \textbf{Hand Tuned} & \textbf{Algorithm} & \textbf{Parameters} & \textbf{Hand Tuned} \\ \hline
        \textbf{SAC} & \makecell{Replay Buffer Size \\ Target Update $\tau$ \\ Target Update Period \\ $\gamma$ \\ Critic Learning Rate \\ Actor Learning Rate \\ Alpha Learning Rate} & \makecell{256 \\ 0.005 \\ 1 \\ 0.99 \\ 0.0003 \\ 0.0003 \\ 0.0003} & \textbf{PPO} & \makecell{Normalize Observations \\ Normalize Rewards \\ Episodes per Iteration \\ Learning Rate \\ Entropy Regularization \\ Importance Ratio Clipping \\ KL Cutoff Factor \\ KL Cutoff Coefficient \\ Initial Adaptive KL Beta \\ Adaptive KL Target \\ Discount Factor} & \makecell{True \\ True \\ See Table \ref{tab:environments} \\ 0.0001 \\ 0.0 \\ 0.0 \\ 2.0 \\ 100 \\ 1.0 \\ 0.01 \\ 0.995} \\ \hline \hline
    \end{tabular}
\end{table}
\setlength{\tabcolsep}{6pt}

\begin{figure*}[t]
	\begin{center}
		\begin{tabular}{cc}
            \subfloat[Training on Humanoid Standup]{\includegraphics[trim=0mm 0mm 0mm 0.35mm,clip,height=5.5cm,keepaspectratio=true]{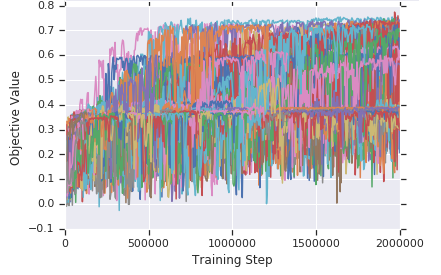}}
    		&
            \subfloat[Histogram of Best Values]{\includegraphics[trim=0mm 0mm 0mm 0mm,clip,height=5.5cm,keepaspectratio=true]{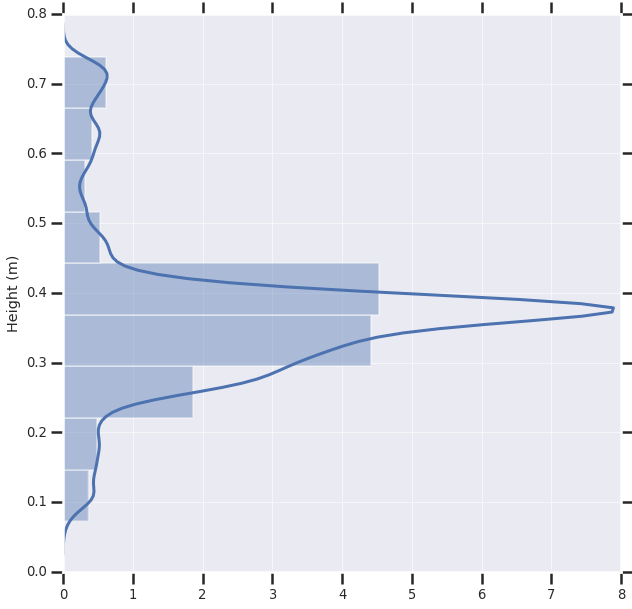}}
		\end{tabular}
		\caption{\small Optimizing Humanoid Standup over a dense true objective. a) Training performance for approximately 100 agents.  b) Histogram of final values achieved by 1000 studies. \label{fig:why_dense_objectives}}
	\end{center}
\end{figure*}

\section{True Objective Selection}
\label{sec:sparse}
In our prior work for point-to-point navigation tasks (\cite{autorl}), we used sparse true objectives such as reaching a goal. For continuous control tasks, however, this is problematic, because sparse true objectives are difficult to achieve, while intermediate stages in learning are valuable. Figure \ref{fig:why_dense_objectives} illustrates this for Humanoid Standup. A sparse true objective for this task is for the agent to stand up to approximately \dist{1.4}, but few agents successfully achieve this height. Instead, plateaus of performance can be seen at intermediate heights of \dist{0.4}, \dist{0.6} and \dist{0.75} where agents have likely learned important intermediate behaviors, such as sitting up or rising to one knee. All of these behaviors look the same to a sparse true objective: they are failures.

\subsection{Single-Task Objectives}
\label{ref:metric-objectives}
Metric-based objectives provide a way out of the sparse optimization issue by splitting the difference between a true sparse objective and a hand-engineered reward. Without committing to reward component weights, one can say that policies that achieve more height are preferred for standup tasks, and policies that achieve more distance are preferred for locomotion. Metric-based single-task objectives \eqref{eq:obj} form a metric space over the space of policies, and induce a partial ordering of all policies \wrt the $<$ relation.  They reach the true objective in the limit and are continuous, providing a clear signal to the evolutionary optimization process. We use two metric-based task objectives: distance traveled for Ant, Walker and Humanoid, and height achieved for Humanoid Standup.

\subsection{Multi-Objective Standard Returns}
\label{ref:standard-objectives}
Another kind of dense true objective is the standard return by which tasks are normally evaluated. The rewards that these returns are based on are generally multiple objectives combined with hand-engineered weights, and may not have the same gradient or maxima as the true value function given the environment dynamics. A parameterization which more closely matches the true value function will encourage agents to make good decisions earlier in training and to explore more fruitful parts of the search space. Therefore, optimizing a reparameterization of the standard reward against the standard return can yield improved performance.

\subsection{Single-Task Objective Evaluation}
Analysis of the videos\footnote{\url{https://youtu.be/svdaOFfQyC8}} for optimization over task objectives reveals that SAC generally outperforms PPO, and AutoRL outperforms hand-tuning and hyperparameter tuning. Hand and hyperparameter tuning frequently fell on PPO, whereas AutoRL for SAC had the fastest travel. For HumanoidStandup, SAC again performed better than PPO, while AutoRL performed better than hand tuning and hyperparameter tuning. No policy fully stood, but both hyperparameter tuning and AutoRL on SAC rose to a consistent crouch.

\setlength{\tabcolsep}{5pt}
\begin{table}[tb]
\scriptsize
    \caption{AutoRL Parameterized Reward Weights with SAC}
	\label{tab:parametrized_rewards_sac}
    \centering
    \begin{tabular}{r|cc|cc|cc|cc}
        \hline \hline
        \textbf{Parameters} & \multicolumn{2}{c}{\textbf{Ant}}& \multicolumn{2}{c}{\textbf{Walker}} & \multicolumn{2}{c}{\textbf{Standup}} & \multicolumn{2}{c}{\textbf{Humanoid}} \\ \hline
        & Std. Obj. & Task Obj. & Std. Obj. & Task Obj. & Std. Obj. & Task Obj. & Std. Obj. & Task Obj. \\ \cline{2-9} 
        \textbf{Achievement} & 0.10205 & 0.34949 & 0.19719 & 0.46260 & 0.33561 & 0.21251 & 0.86552 & 0.90872  \\
        \textbf{Cost} & 0.34042 & 0.15624 & 0.77295 & 0.97506 & 0.65550 & 0.96195 & 0.54985 & 0.99147 \\
        \textbf{Impact} & 0.20531 & 0.53880 & N/A & N/A & 0.34548 & 0.89288 & 0.31891 & 0.33972 \\
        \textbf{Survival} & 0.05205 & 0.01969 & 0.01421 & 0.47166 & 0.86399 & 0.97217 & 0.02702 & 0.06027 \\ \hline \hline
    \end{tabular}
\end{table}
\setlength{\tabcolsep}{6pt}

\setlength{\tabcolsep}{5pt}
\begin{table}[tb]
\scriptsize
    \caption{AutoRL Parameterized Reward Weights with PPO}
	\label{tab:parametrized_rewards_ppo}
    \centering
    \begin{tabular}{r|cc|cc|cc|cc}
        \hline \hline
        \textbf{Parameters} & \multicolumn{2}{c}{\textbf{Ant}}& \multicolumn{2}{c}{\textbf{Walker}} & \multicolumn{2}{c}{\textbf{Standup}} & \multicolumn{2}{c}{\textbf{Humanoid}} \\ \hline
        & Std. Obj. & Task Obj. & Std. Obj. & Task Obj. & Std. Obj. & Task Obj. & Std. Obj. & Task Obj. \\ \cline{2-9}
        \textbf{Achievement} & 0.73208 & 0.97706 & 0.08453 & 0.71359 & 0.53268 & 0.67899 & 0.40621 & 0.19427  \\
        \textbf{Cost} & 0.84979 & 0.73720 & 0.21984 & 0.04812 & 0.73103 & 0.88079 & 0.32404 & 0.49128 \\
        \textbf{Impact} & 0.35485 & 0.80661 & N/A & N/A & 0.62980 & 0.40959 & 0.35179 & 0.84411 \\
        \textbf{Survival} & 0.46004 & 0.44921 & 0.00677 & 0.06200 & 0.50112 & 0.01883 & 0.08046 & 0.046723 \\ \hline \hline
    \end{tabular}
\end{table}
\setlength{\tabcolsep}{6pt}

\subsection{Multi-Objective Standard Return Evaluation}
Evaluation of optimization over the standard return was similar to evaluation over the task objective, except we collected returns over the complete standard rewards listed in Table \ref{tab:tuned_rewards} (see Section \ref{ref:standard-objectives}). AutoRL was consistently superior for SAC and superior for Walker and Humanoid in PPO.

Analysis of the videos\footnote{\url{https://youtu.be/svdaOFfQyC8}} for optimization over the standard returns similarly reveals that SAC generally outperforms PPO, that AutoRL outperforms hand-tuning and hyperparameter tuning, and that hand and hyperparameter tuning frequently fell in PPO, whereas AutoRL for SAC had the fastest travel. For HumanoidStandup, SAC again performed better than PPO, while AutoRL performed better than hand tuning and hyperparameter tuning. AutoRL on SAC was the only policy that fully stood, though it got into a falling and standing loop, whereas hyperparameter tuning rose to a consistent crouch.

\subsection{Cross-Evaluation of Single-Task Objectives and Standard Returns}
\label{sec:cross-eval}

All the environments in Table \ref{tab:environments} define standard rewards with predefined components listed in Table \ref{tab:tuned_rewards}. AutoRL's reward optimization changes the parameterization of these components, but the return collected over the standard reward parameterization is the normal way that these environments are evaluated. However, task objective optimization is evaluating over a different objective - normally, just the achievement objective of Table \ref{tab:tuned_rewards}. We would expect optimizing over a different reward to produce different performance on the standard return; conversely, we would expect the standard return to produce different performance on the task objectives.

\begin{figure*}[t]
	\begin{center}
		\begin{tabular}{ccccc}
		&\small Ant-v1 &\small  Walker2D-v1 &\small HumanoidStandup-v1 & \small Humanoid-v1 \\
            \multirow{2}{*}{\rotatebox{90}{\small Task Obj. on Std. Rwds.}}
			&
			\subfloat[]{\includegraphics[trim=2mm 2mm 2mm
			2.2mm,clip,width=0.22\textwidth,keepaspectratio=true]{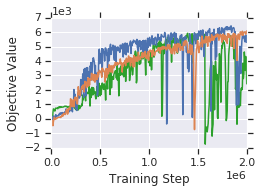}\label{fig:sac_ant_dg_avgR}} 
			&
			\subfloat[]{\includegraphics[trim=2mm 2mm 2mm 2mm
			2.2mm,clip,width=0.22\textwidth,keepaspectratio=true]{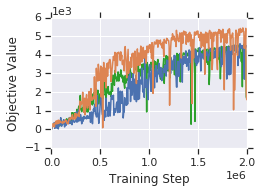}\label{fig:sac_walker2d_dg_avgR}} 
			&
			\subfloat[]{\includegraphics[trim=2mm 2mm 2mm
			2mm,clip,width=0.22\textwidth,keepaspectratio=true]{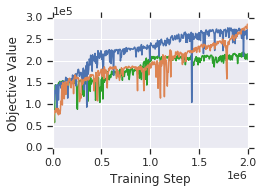}\label{fig:sac_standup_dg_avgR}} 
			&
			\subfloat[]{\includegraphics[trim=2mm 2mm 2mm
			2.2mm,clip,width=0.22\textwidth,keepaspectratio=true]{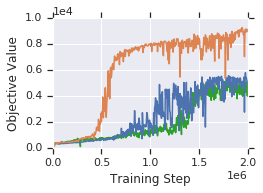}\label{fig:sac_humanoid_dg_avgR}}
            \\
            &
            \subfloat[]{\includegraphics[trim=2mm 2mm 2mm
			2.2mm,clip,width=0.22\textwidth,keepaspectratio=true]{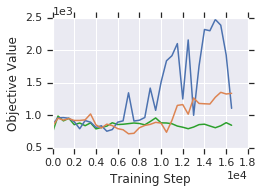}\label{fig:ppo_ant_dg_avgR}} 
            &
            \subfloat[]{\includegraphics[trim=2mm 2mm 2mm
			2.2mm,clip,width=0.22\textwidth,keepaspectratio=true]{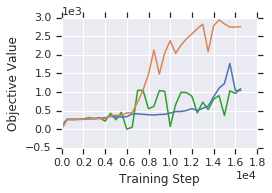}\label{fig:ppo_walker2d_dg_avgR}} 
            &
            \subfloat[]{\includegraphics[trim=2mm 2mm 2mm
			2.2mm,clip,width=0.22\textwidth,keepaspectratio=true]{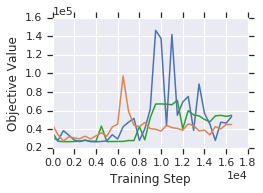}\label{fig:ppo_standup_dg_avgR}} 
            &
            \subfloat[]{\includegraphics[trim=2mm 2mm 2mm
			2.2mm,clip,width=0.22\textwidth,keepaspectratio=true]{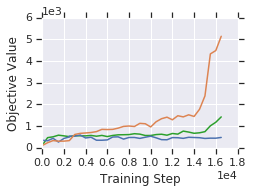}\label{fig:ppo_humanoid_dg_avgR}}
            \\
            \multirow{2}{*}{\rotatebox{90}{\small Std. Obj. on Task Rwds.}}
			&
			\subfloat[]{\includegraphics[trim=2mm 2mm 2mm
			2.2mm,clip,width=0.22\textwidth,keepaspectratio=true]{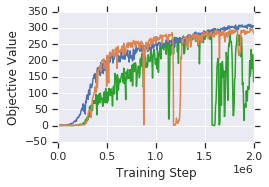}\label{fig:sac_ant_sr_avgDG}} 
			&
			\subfloat[]{\includegraphics[trim=2mm 2mm 2mm
			2.2mm,clip,width=0.22\textwidth,keepaspectratio=true]{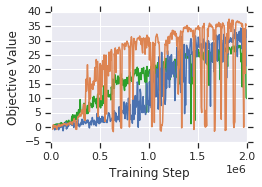}\label{fig:sac_walker2d_sr_avgDG}} 
			&
			\subfloat[]{\includegraphics[trim=2mm 2mm 2mm
			2.2mm,clip,width=0.22\textwidth,keepaspectratio=true]{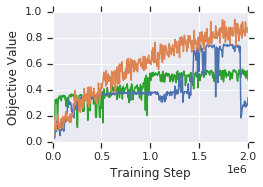}\label{fig:sac_standup_sr_avgDG}} 
			&
			\subfloat[]{\includegraphics[trim=2mm 2mm 2mm
			2.2mm,clip,width=0.22\textwidth,keepaspectratio=true]{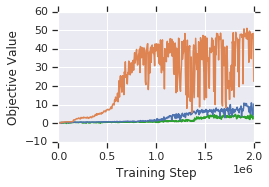}\label{fig:sac_humanoid_sr_avgDG}}
            \\
            &
            \subfloat[]{\includegraphics[trim=2mm 2mm 2mm
			2.2mm,clip,width=0.22\textwidth,keepaspectratio=true]{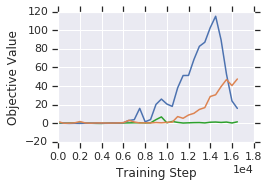}\label{fig:ppo_ant_sr_avgDG}} 
            &
            \subfloat[]{\includegraphics[trim=2mm 2mm 2mm
			2.2mm,clip,width=0.22\textwidth,keepaspectratio=true]{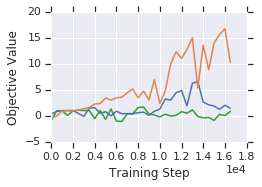}\label{fig:ppo_walker2d_sr_avgDG}} 
            &
            \subfloat[]{\includegraphics[trim=2mm 2mm 2mm
			2.2mm,clip,width=0.22\textwidth,keepaspectratio=true]{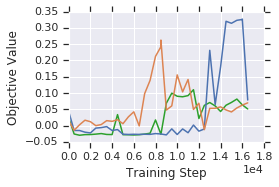}\label{fig:ppo_standup_sr_avgDG}} 
            &
            \subfloat[]{\includegraphics[trim=2mm 2mm 2mm
			2.2mm,clip,width=0.22\textwidth,keepaspectratio=true]{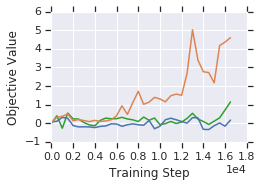}\label{fig:ppo_humanoid_sr_avgDG}} 
		\end{tabular}
		\begin{tabular}{cccc}
        \hspace{5mm} &\xdash[green] Hand Tuned & \xdash[blue] Hyperparameter Tuned & \xdash[orange] Reward Tuned \textbf{(ours)}
        \end{tabular}
		\caption{\small AutoRL policies optimized over task objectives evaluated on standard return a-d) using SAC and e-h) using PPO. AutoRL policies optimized over standard return evaluated on task objectives i-l) using SAC and m-p) using PPO. \label{fig:meta_objective_comparison_learning_curves}}
	\end{center}
\end{figure*}

To enable a fair comparison of both conditions, we conducted a cross-evaluation study in which we evaluated policies optimized for task objectives against the standard returns (Fig. \ref{fig:meta_objective_comparison_learning_curves} a-h), as well as policies optimized on the standard returns against the  task objectives (Fig. \ref{fig:meta_objective_comparison_learning_curves} i-p). These results show task objectives and standard returns have similar performance, suggesting task objectives obtain good policies with reduced reward engineering effort.

As discussed in the text, detailed analysis reveals differences in these objectives. Videos show differences in style, and in Humanoid, the task objective agents travel the farthest for both PPO and SAC (Figures \ref{fig:meta_objective_comparison}a and \ref{fig:meta_objective_comparison}b dark red vs light red), while on other tasks optimizing over the task objective is comparable to optimizing over the standard reward.  Unsurprisingly, AutoRL over the standard reward produces the highest scores when evaluated on the standard reward in most conditions.

\end{document}